\documentclass{bmcart}
\usepackage[utf8]{inputenc} 

\usepackage{listings}
\usepackage{xcolor}

\colorlet{punct}{red!60!black}
\definecolor{background}{HTML}{EEEEEE}
\definecolor{delim}{RGB}{20,105,176}
\colorlet{numb}{magenta!60!black}

\lstdefinelanguage{json}{
    basicstyle=\small\ttfamily,
    numbers=left,
    numberstyle=\scriptsize,
    stepnumber=1,
    numbersep=8pt,
    showstringspaces=false,
    breaklines=true,
    frame=lines,
    backgroundcolor=\color{background},
    literate=
     *{:}{{{\color{punct}{:}}}}{1}
      {,}{{{\color{punct}{,}}}}{1}
      {\{}{{{\color{delim}{\{}}}}{1}
      {\}}{{{\color{delim}{\}}}}}{1}
      {[}{{{\color{delim}{[}}}}{1}
      {]}{{{\color{delim}{]}}}}{1},
}

\startlocaldefs
\usepackage{url}
\usepackage{todonotes}
\usepackage[utf8]{inputenc}

\usepackage{tikz}
\usepackage{pgfplots}
\usepgfplotslibrary{dateplot}
\usepgfplotslibrary{statistics}
\pgfplotsset{compat=1.11} 

\usetikzlibrary{shapes,arrows}
\usepackage{scrextend}
\usepackage[hidelinks]{hyperref}
\usepackage{subfig}

\usepackage{float}
\usepackage{filecontents}
\usepackage{tikzscale}

\usepackage{listings}

\endlocaldefs

\begin{document}

\begin{frontmatter}

\begin{fmbox}
\dochead{Research}


\title{SIA: A Scalable Interoperable Annotation Server for Biomedical Named Entities}


\author[
   addressref={aff1},                   
   corref={aff1},                       
   noteref={n1}                        
]{\inits{JK}\fnm{Johannes} \snm{Kirschnick}}
\author[
   addressref={aff1},
   noteref={n1}
]{\inits{PT}\fnm{Philippe} \snm{Thomas}}
\author[
   addressref={aff1},
]{\inits{RR}\fnm{Roland} \snm{Roller}}
\author[
   addressref={aff1},
   email={leonhard.hennig@dfki.de}
]{\inits{LH}\fnm{Leonhard} \snm{Hennig}}

\address[id=aff1]{
  \orgname{DFKI Language Technology Lab}, 
  \street{Alt-Moabit 91c},                     %
  \city{Berlin},                              
  \cny{Germany}                                    
}


\begin{artnotes}
\note[id=n1]{Equal contributor} 
\end{artnotes}

jkirschnick@gmail.com, philippe.thomas@dfki.de, roland.roller@dfki.de, leonhard.hennig@dfki.de

\end{fmbox}


\begin{abstractbox}

\begin{abstract} 
Recent years showed a strong increase in biomedical sciences and an inherent increase in publication volume.
Extraction of specific information from these sources requires highly sophisticated text mining  and information extraction tools.
However, the integration of freely available tools into customized workflows is often cumbersome and difficult. 
We describe SIA (Scalable Interoperable Annotation Server), our contribution to the BeCalm-Technical interoperability and performance of annotation servers (BeCalm-TIPS) task, a scalable, extensible, and robust annotation service.
The system currently covers six named entity types (\emph{i.e.,} Chemicals, Diseases, Genes, miRNA, Mutations, and Organisms)
and is freely available under Apache 2.0 license at \url{https://github.com/Erechtheus/sia}.

\end{abstract}



\end{abstractbox}
%

\end{frontmatter}



\section{Introduction}
A vast amount of information on biomedical processes is scattered over millions of scientific publications.
Manual curation of this information is expensive and cannot keep up with the ever increasing volume of biomedical literature~\cite{Hunter06}. 
To this end, several sophisticated natural language processing tools have been proposed to assist professionals in finding specific information from texts.
Many of these highly specialized tools are provided as open source projects to the community.
However, the integration of state-of-the-art open source extractors into customized text-mining workflows is often difficult and cumbersome~\cite{Rheinlander2016,thomas2013a}. 
Standardized interchange formats, such as BioC~\cite{Comeau2013}, enable the exchange of text mining results but the initial set-up of these tools is still an unsolved issue.
Exposing tools via public web services implementing common specifications bypasses this problem and allows a code-agnostic integration of specific tools by providing an interoperable interface to third parties.
This enables simple integration, comparison, and aggregation of different state-of-the-art tools.
In this publication we present SIA, our contribution to the BeCalm-TIPS task~\cite{Perez2017}. SIA is a robust, scalable, extensible, and generic framework to combine multiple named entity recognition tools into a single system.

The publication is organized as follows: First, we briefly introduce the BeCalm-TIPS task and its requirements. We then give an overview of the SIA system architecture, followed by a detailed description of the implementation and the error handling features. This is followed by a scalability experiment conducted on a large dump of PubMed articles and a discussion of the results. We end with a summary and a future work section. 

\section{BeCalm-TIPS Task Overview}
The following section provides a short introduction to the BeCalm-TIPS Task, focusing on the payloads annotation servers had to accept and respond with. A detailed description of the task is available in~\cite{Perez2017}.

The task set out to define a testbed for comparing different annotation tools by making them accessible via public web endpoints which exchange standardized JSON messages.
It required participants to register their endpoint and a set of supported named entity types with a system managed by the task organizers. Over the course of the task, this endpoint received a number of annotation requests. Each request was not required to be processed interactively, just the message reception had to be acknowledged. Once the annotations were generated by the annotation server, they had to be sent back to a dedicated endpoint - via a separate HTTP request.

\begin{lstlisting}[language=json,firstnumber=1,caption=JSON payload excerpt for an annotation request, label=exampleRequest]
{"documents":
   [{"document_id": "BC1403854C", "source":"PUBMED"}, ...],
 "types": ["DISEASE", "MUTATION", "MIRNA"],
 "communication_id": 1581}
\end{lstlisting}

Listing \ref{exampleRequest} shows an excerpt of the JSON payload for an annotation request. It consists of a list of document identifiers and their respective source. As no text was transmitted, participants where required to implement their own text retrieval component to fetch the title, abstract and potentially full text for each document prior to processing. A type field specified the list of named entities to be identified. A unique communication identifier was passed along, which had to be included in any outgoing messages in order to correlate individual requests and responses.

\begin{lstlisting}[language=json,firstnumber=1,caption=JSON payload excerpt for an annotation response, label=exampleResponse]
[{"document_id":"BC1403855C", "section":"A",
  "init":410,   "end":419,    "score":1.0,
  "type":"DISEASE", "annotated_text":"periosteum" }, ...]
\end{lstlisting}

Once the annotation server acknowledged the reception of a request it had a specified amount of time to respond. Listing \ref{exampleResponse} shows a snippet of such a response. It contains a list of detected annotations across all requested documents, identifying the text source section (abstract \textit{A} or title \textit{T}), the start and end positions within it, a confidence score, and the extracted named entity type as well as the annotated text itself.

The task merely specified the required input- as well as output schemata and gave participants full control over the implementation of their system as well as which annotation types they wanted to support.

\section{SIA - General Architecture}
This section describes the architecture of SIA, our contribution to the BeCalm-TIPS Task. Figure~\ref{figure:architecture} shows a high level overview of the general architecture, which was designed around the following three main goals:
\begin{enumerate}
\item \textbf{Scalability} The ability to handle large amounts of concurrent requests, tolerating bursts of high request rates over short periods of time.
\item \textbf{Robustness} Temporary failures (\emph{e.g., } networking problems or server failures) should be handled transparently and not lead to dropped requests. 
\item \textbf{Extensibility} Enable simple integration of arbitrary NLP tools to reduce initial burden for providing an annotation service. 
\end{enumerate}

To achieve these goals, SIA is split into three components, the \textbf{front end}, \textbf{back end}, and \textbf{result handling}, respectively. The front end handles the interactive aspects of the system, while the other components implement the system's non-interactive elements. 

To connect these components, we opted for a message based architecture, which links each component to a central message bus, over which they exchange messages. Incoming annotation requests are translated into messages by the front end, and subsequently processed by the back end. Once processing is finished the final result is handled by the result handler. To this end, SIA defines a configurable message flow for each message, which incorporates fetching raw texts, running a set of annotators, aggregating the results and forwarding them to a result handler. The configuration defines the actual processing steps, the set of annotator components to use, which document fetchers to enable and how to deal with the results. For example, a processing flow could fetch PubMed articles from a public endpoint, handle all requests for \textit{Mutations} with the SETH~\cite{Thomas2016a} tagger and send annotation results back to the requester. The overall processing flow is expressed as an ordered sequence of message transformation and aggregation steps, while the configuration allows to extend the actual processing flow with new annotator and document fetcher components. Interested readers are referred to Enterprise Integration Patterns~\cite{Hohpe:2003:EIP:940308} for a detailed discussion of the different message handling and transformation strategies that SIA employs.

To handle messages, persistent queues are defined as input and output buffers for all components, where a subsequent component consumes from the previous component's output queue. These queues are stored for the entirety of the system's lifetime. 
This architecture further provides fault tolerant and scalable processing. Fault tolerance is enabled through component wise acknowledgment of each successful message processing, which allows replaying all unacknowledged messages during system recovery, while scalability is achieved through component replication and round robin based message forwarding for increased message throughput. 

Messages, the data objects in our architecture, carry information through the system and are composed of a \textsc{Header} and \textsc{Payload} part. The \textsc{Header} contains meta information, such as expiry date, global ids and requested annotation types, and is used by the system to route messages to the respective consumers. The \textsc{Payload} contains the actual data to be processed. 

\section{Implementation Details}
SIA is implemented in Java and uses RabbitMQ~\cite{rabbitMQ} 
as its message bus implementation. 
In the following each individual component of SIA is described in detail.

\subsection{Front end}
The front end encapsulates the annotation processing for the clients and serves as the entry point to the system. Currently it provides a REST endpoint according to the Becalm-TIPS task specification. Other entry points, such as interactive parsing can easily be added.
Incoming requests are translated into messages and forwarded to an input queue. This way, the overall processing in the front end is very lightweight and new requests can be handled irrespectively of any ongoing annotation processing. Furthermore, the back end does not need to be online at the time of a request, but instead could be started dynamically based on observed load.

To handle multiple concurrent requests with varying deadlines, we make use of the fact that the input queue is a priority queue, and prioritize messages with an earlier expiry date. Already running requests will not be canceled, the priority is just used as a fast path to the front of the queue. 
The message expiry date, as provided by the calling clients, is translated into a message priority using the currently processed messages and their deadlines as well as past elapsed processing time statistics to estimate the individual message urgency. 

The front end also handles validation and authorization, which moves this logic into a central place.
Furthermore, the front end provides a monitoring entry point into the system, reporting computation statistics, such as request rates, recent document types as well as back end processing counters, for display in dashboards and for observing the current health of the system.

\subsection{Back end}

The back end is concerned with fetching documents from the supported corpus providers, calling the requested annotators for each resulting text fragment, aggregating the results and feeding them to a result handler.
It is modeled using a pipeline of message transformations, which subsequently read from message queues and post back to new ones. The message flow starts by reading new requests from the input queue, which is filled by the front end. The front end does not communicate directly with the back end, but instead the input queue is used as a hand over point. Since a single annotation request, in the case of the Becalm-TIPS task specification, may contain multiple document ids, incoming messages are first split into document-level messages. Splitting takes one message as input and generates as many individual messages as there are document ids specified. The raw text for each document is then retrieved by passing the messages through corpus adapters. The outcome is the retrieved text, separated into fields for abstract, title and potentially full text.

Raw texts messages are then delivered to registered annotators using a scatter-gather approach. 
Each message is duplicated (scattered) to the respective input queue of a qualified annotator. To find the annotator, the required annotator type per message is translated into a queue name, as each annotator has a dedicated input queue. Upon completion all resulting annotation messages are combined together (gathered) into a single message.
This design allows to add new annotators by registering a new input queue and adding it to the annotation type mapping. This mapping is also exposed as a runtime configuration, which allows to dynamically (de-)activate annotators.

The next step in the message flow aggregates all annotation results across all documents that belong to the same request. It is the inverse of the initial split operation, and aggregates all messages sharing the same unique request id into a single one.
Overlapping annotations (\emph{e.g., }from different annotator components) are merged without any specific post processing. This strategy allows end users the highest flexibility as annotations are not silently modified. 
Finally, the aggregated message is forwarded to the output queue.

While the processing flow is specified in a sequential manner, this does not entail single threaded execution. Each individual transformer, such as a corpus adapter or an annotator, works independently and can be further scaled out, if they present a processing bottleneck.
Furthermore, multiple requests can be handled in parallel at different stages of the pipeline. Transacting the message delivery to each transformer and retrying on failure, provides the fault tolerance of the system.
Overall, the back end specifies a pipeline of an ordered execution flow and provides two injection points where users, through configuration, can add new functionality with additional corpus adapters or new annotation handlers.

To increase the throughput of the back end, multiple instances of SIA can be started on different machines, where each instance would process requests in a round robin fashion. 

\subsubsection{Supported Annotators}
To illustrate the extensibility of our approach, we integrated Named Entity Recognition (NER) components for six different entity types into SIA:
Mutation names are extracted using SETH~\cite{Thomas2016a}.
For micro-RNA mentions we implement a set of regular expressions~\cite{mirNer},
which follow the recommendations for micro-RNA nomenclature~\cite{Ambros2003}.
Disease names are recognized using a dictionary lookup~\cite{Aho1975}, generated from UMLS disease terms~\cite{Bodenreider04}, and by using the DNorm tagger~\cite{leaman2013dnorm}. Chemical name mentions are detected with ChemSpot~\cite{Rocktaschel2012}, Organisms using Linnaues~\cite{Gerner2010} and Gene mentions using Banner~\cite{Leaman2008}.

Listing \ref{lst:annotator} shows the general interface contract SIA is expecting for each annotator. Each annotator receives an input text and is simply expected to return a set of found annotations. Thus integrating any of the aforementioned annotators, as well as new ones, is as simple as implementing this interface and registering a new queue mapping. 

Annotation handlers can be hosted inside of SIA, within the same process, or externally, in a separate process. External hosting allows to integrate annotation tools across programming languages, operating systems and servers. This is especially useful since most annotators have conflicting dependencies that are either very hard or impossible to resolve. For example, ChemSpot and DNorm use different versions of the Banner tagger which make them candidates for external hosting. Multiple servers can also be used to increase the available resources for SIA, \emph{e.g., } when hosting all annotators on the same machine exceeds the amount of available memory.

\subsubsection{Corpus Adapters}
SIA contains corpus adapters for PubMed, PMC, and the BeCalm patent- and abstract servers, which communicate to external network services. These components are represented as transformers, which process document ids and return retrieved source texts. They are implemented following the interface definition shown in Listing \ref{lst:fetcher}. If an adapter supports bulk fetching of multiple documents, we feed a configurable number of ids in one invocation.

As retrieving the full text translates into calling a potentially unreliable remote service over the network, retry on failure is used in case of recoverable errors. This is backed up by the observation that the most commonly observed error was a temporarily unavailable service endpoint. To spread retries, we use exponential backoff on continuous failures with an exponentially increasing time interval, capped at a maximum (initial wait $1s$, multiplier $2$, max wait $60s$). If a corpus adapter fails to produce a result after retries are exhausted, we mark that document as unavailable and treat it as one without any text. This allows a trade-off between never advancing the processing, as a document could be part of a set of documents to be annotated, and giving up too early in case of transient errors.

\subsection{Result handler}
The result handler processes the aggregated annotation results from the back end, by consuming from a dedicated output queue. 
We implemented a REST component according to the TIPS task specification, which posts these annotations back to a dedicated endpoint. Additional handlers, such as statistics gatherer or result archiver, can easily be added.

\lstset{language=Java}
\begin{lstlisting}[frame=lines, float=*, caption=Interface definition for SIA annotators, label=lst:annotator]
public interface Annotator {
    Set<PredictionResult> annotate(InputText payload);
}
\end{lstlisting}

\begin{lstlisting}[frame=lines, float=*, caption=Interface definition for SIA corpus adapters, label=lst:fetcher]
public interface CorpusAdapter {
    InputText load(String documentID);
}
\end{lstlisting}

\section{Failure Handling}
In the following we describe the failure handling strategies across the different components within SIA.

\noindent \textbf{Invalid requests}
Client calls with wrong or missing information are handled in the front end using request validation.
Such invalid requests are communicated back to the caller with detailed error descriptions.

\noindent \textbf{Backpressure}
To avoid that a large number of simultaneous requests can temporarily overload the processing system, SIA buffers all accepted requests in the input queue - using priorities to represent deadlines. 

\noindent \textbf{Front end fails}
If the front end stops, new requests are simply not accepted, irrespective of any ongoing processing in the back end.

\noindent \textbf{Back end unavailable} Messages are still accepted and buffered when there is enough storage space, otherwise the front end denies any new annotation requests.

\noindent \textbf{Back end fails} If the back end stops while there are still messages being processed, these are not lost but retried upon restart. This is enabled by acknowledging each message only upon successful processing per component. 

\noindent \textbf{Corpus adapter fails} Each adapter retries, using exponential backoff, to fetch a document before it is marked as unavailable. As the BeCalm-TIPS task does not specify how to signal unavailable documents, these are just internally logged. Any subsequent processing treats a missing document as one with no content. 

\noindent \textbf{Annotator fails} If an annotator fails on a particular message, this can potentially harm the entire back end when annotators are embedded in the system.
As annotators are software components not under the control of the processing pipeline, we catch all recoverable errors and return zero found annotations in these cases - logging the errors for later analysis.

\noindent \textbf{Result Handling fails} The BeCalm-TIPS task description expects the result of an annotation request to be delivered to a known endpoint. If this fails, the delivery is retried in a similar manner to the corpus adapter failure handling.

\noindent \textbf{Message expired} Clients can define a deadline for results. This is mapped to a time-to-live attribute of each message. This results in automatically dropping any expired messages from the message bus.

\section{Performance Test}

To test the scalability as well as extensibility of SIA we performed an offline evaluation, focusing on throughput. To this end we extended the front end to accept full text documents and added an identity corpus adapter which simply returns the document text from the request message itself. Furthermore, we added a result handler, which writes all results into a local file. By adding these components, we turned SIA into an offline annotation tool, that can be fed from a local collection of text documents without relying on external document providers.

For the test, we used a dump of $207.551$ PubMed articles\footnote{Using files 922 to 928 from~\cite{NCBI}} and enabled all internal annotators (SETH, mirNer, Linnaues, Banner, DiseaseNer) in a single SIA instance, as well as ChemSpot using the external integration on the same machine. The experiment was run on a Server with 2 Intel Xeon E5-2630 processor (8 threads, 16 cores each) and 256 GB RAM running Ubuntu 12.04. To simulate the scaling behavior, we varied the degree of parallelism used by SIA from $1$ to $5$ respectively and measured the overall time to annotate all documents. The parallelism controls the number of messages consumed from the input queue simultanously. Table \ref{table:scalability} shows the resulting runtimes. When increasing the parallelism we see a decrease of processing times with a speedup of up to $3\times$ compared to single threaded execution. Increasing the parallelism further did not yield lower processing times, as the processing is mainly CPU bound, with a ceiling hit with $5$ parallel threads. This highlights that SIA is fully capable of exploiting all available CPU resources, achieving a throughput of more than $70$ documents per second. Using the parallelism within SIA furthermore enables to effortlessly provide parallel processing for exiting annotators that are otherwise hard to scale.

\section{Discussion}

SIA itself is very lightweight and runs anywhere given a Java environment and a connection to RabbitMQ. Annotators can be directly embedded or configured to run externally, exchanging messages through the bus. During the BeCalm-TIPS tasks, we deployed SIA into Cloud Foundry, a platform as a service provider, which enables deployments of cloud containers \cite{Kirschnick:2012:TAD:2275356.2275364}. The front- and back end with embedded result handling were deployed as two separate application containers connected to a hosted instance of RabbitMQ.  To limit the resource consumption, we only enabled the SETH, mirNER and DiseaseNER annotators.

Figure \ref{fig:processing} shows the received and processed annotation requests over the course of a four week period during the task. It highlights that our system is capable of sustaining a high number of daily requests, with more than $14.000$ daily requests received at maximum. Furthermore we observed that the request handling time during these weeks was dominated by individual corpus downloading times, which make up about $50\%$ of the overall processing time. This validates our decision to support bulk downloading of documents, as this amortizes the networking overhead for each document, over a number of documents. Processing each annotation request in total took less than two seconds for the configured annotators. 
We observed higher annotation times for PubMed articles, which is partially due to higher server response times and the need for more sophisticated result parsing. 
We also estimated the message bus overhead to about $10\%$, stemming from individual message serialization and persistence compared to running the annotators stand alone - an acceptable slowdown which is easily compensated by additional parallelism.

\section{Summary and Future Work}
We described SIA, our contribution to the BeCalm-TIPS task, which provides scalability - through component replication, fault tolerance - through message acknowledgement, and extensibility - through well defined injection points -- with a particular emphasis on failure handling.
The message-based architecture proved to be a good design blueprint, 
which can be extended with additional components. To further provide scalable processing, a suggested improvement is to automate the back end scaling by coupling it with an input queue length monitoring. This would allow to scale the back end up or down in response to changes in observed load. 
One interesting further development path is to port SIA to a distributed streaming environment such as Flink~\cite{flink} or Spark~\cite{spark}. These systems reduce the overhead of the message bus at the expense of more complex stream processing and result aggregation. While many of the existing components could be reused, some engineering effort would need to be spent on implementing a fault tolerant aggregation, integrating the potentially unreliable corpus adapters. 

To encourage further discussion, the source of our current solution is freely available under an Apache 2.0 license at \url{https://github.com/Erechtheus/sia}, along with detailed guides on how to run and deploy the system.


\begin{backmatter}



\section*{List of abbreviations}

\begin{itemize}
\item NER -- Named Entity Recognition
\item SIA -- Scalable Interoperable Annotation Server
\item TIPS -- Technical interoperability and performance of annotation servers
\end{itemize}

\section*{Availability and requirements}

\begin{itemize}
\item Project name: SIA: Scalable Interoperable Annotation Server
\item Project home page: https://github.com/Erechtheus/sia
\item Operating system(s): Platform independent
\item Programming language: Java
\item Other requirements:  Java 1.8 or higher
\item License:  Apache License, Version 2.0

\end{itemize}

\section*{Competing interests}
The authors declare that they have no competing interests.

\section*{Author's contributions}
JK and PT equally contributed to the implementation of SIA and to writing the manuscript. LH and RR conducted the scalability experiments, and contributed to the manuscript. All authors read and approved the final manuscript.

\section*{Funding} 
This  research  was  partially  supported  by  the  German  Federal  Ministry  of  Economics  and Energy   (BMWi)   through   the   projects   MACSS (01MD16011F), SD4M (01MD15007B) and by the German Federal Ministry of Education and Research (BMBF) through the project BBDC (01IS14013E).

\section*{Author's information}
Email addresses:

JK: jkirschnick@gmail.com

PT: philippe.thomas@dfki.de

RR: roland.roller@dfki.de

LH: leonhard.hennig@dfki.de

\bibliographystyle{bmc-mathphys} 
\bibliography{bmc_article}      




\section*{Figures}

\begin{figure*}[h!]
\centering
\includegraphics[width=\textwidth]{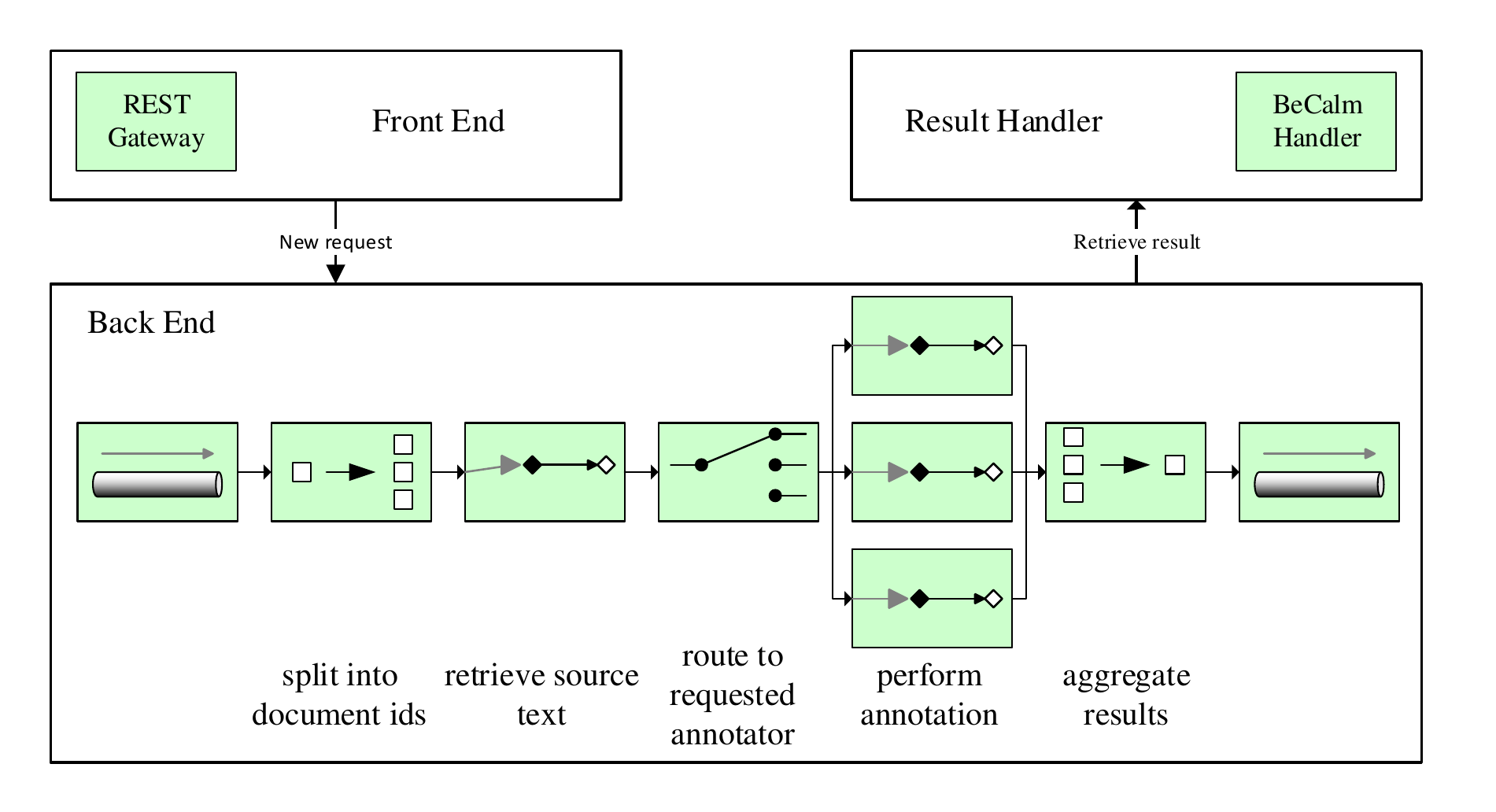}
\captionsetup{width=.92\linewidth}
\caption{\csentence{General Architecture of SIA}. The front end handles new requests and forwards them to the back end over a message bus. Each message is transformed through a series of components, which in turn are connected via named queues. The result handler collects the annotation responses and returns them to the calling client.}
\label{figure:architecture}
\end{figure*}

\begin{figure}[t]
\centering
\subfloat[][Daily request rates]{\resizebox{0.45\textwidth}{!}{
\includegraphics[width=\textwidth]{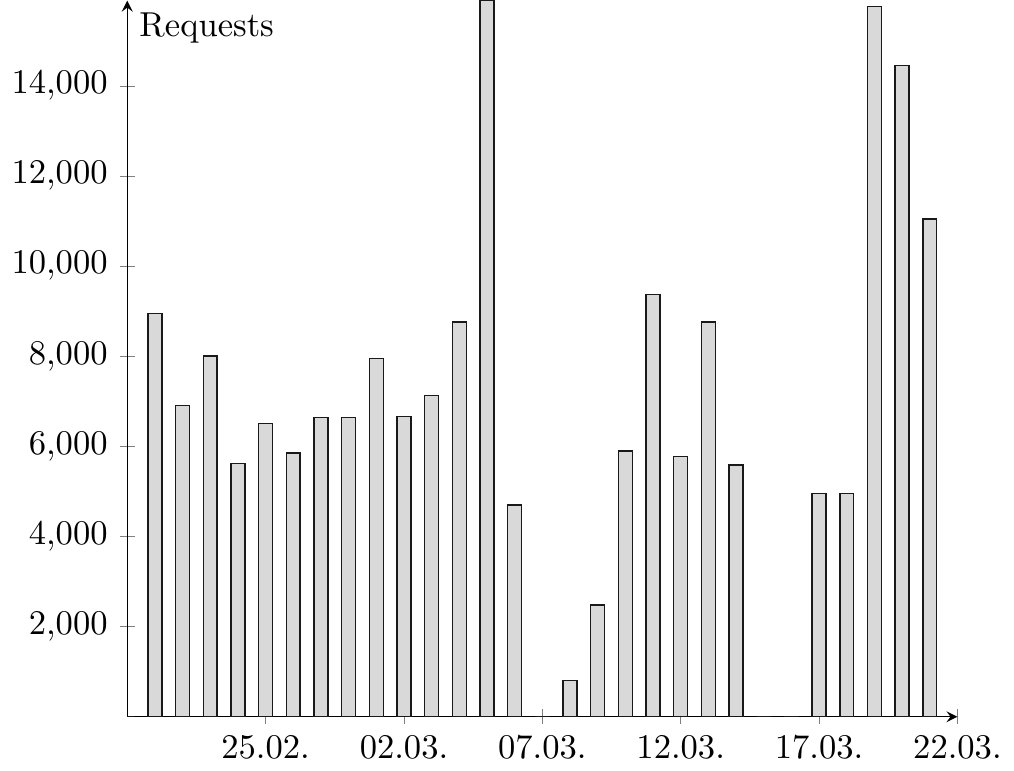}
}}
\subfloat[][Requests processing times for different endpoints]{\resizebox{0.45\textwidth}{!}{
\includegraphics[width=\textwidth]{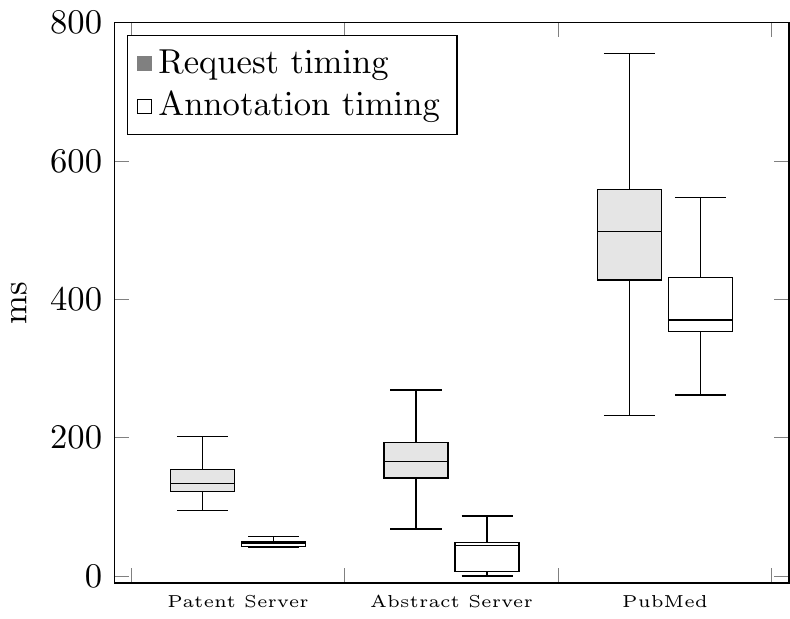}
}}
\captionsetup{width=.92\linewidth}
\caption{Processing statistics over a four week period and request times per corpus, reporting complete processing and annotation timings separately.}
\label{fig:processing}
\end{figure}


\section*{Tables}
\begin{table}[h!]
\begin{tabular}{cccc}
\hline
Parallelism  & Processing time & Throughput & Improvement\\ 
             & (seconds)       & (doc/s) & \\ \hline
$1$ & $8.151$ & 25 & \\
$2$ & $4.551$ & 46 & $1.79 \times$ \\
$3$ & $3.412$ & 61 & $2.39 \times$ \\ 
$4$ & $3.032$ & 68 & $2.69 \times$ \\ 
$5$ & $2.712$ & 77 & $3.01 \times$ \\ \hline
\end{tabular}
\caption{\textbf{Scalability experiment results}. Processing times with varying degree of parallelism, analyzing $207.551$ PubMed articles with all internal annotators (SETH, mirNer, Linnaues, Banner, DiseaseNer) and ChemSpot using a single instance of SIA.}
\label{table:scalability}
\end{table}



\end{backmatter}
\end{document}